\documentclass[10pt,twocolumn,letterpaper]{article}
\usepackage{cvpr}

\usepackage{algorithm}
\usepackage{algorithmic}
\usepackage{listings}
\usepackage{mathtools}

\usepackage{graphicx}
\usepackage{amsmath}
\usepackage{amssymb}
\usepackage{arydshln}
\usepackage{booktabs}
\usepackage{bbding}
\usepackage{enumitem}
\usepackage[table,xcdraw,dvipsnames]{xcolor}
\usepackage{multirow}
\usepackage{amsthm}
\usepackage{mathrsfs}
\usepackage{diagbox}
\usepackage{bm}
\usepackage{newfloat}
\usepackage{xspace}
\usepackage{xcolor}
\usepackage[table]{xcolor}
\definecolor{citecolor}{RGB}{0, 113, 188}
\usepackage[pagebackref,breaklinks,colorlinks, citecolor=citecolor]{hyperref}
\usepackage{makecell}
\usepackage[capitalize]{cleveref}
\crefname{section}{Sec.}{Secs.}
\Crefname{section}{Section}{Sections}
\Crefname{table}{Table}{Tables}
\crefname{table}{Tab.}{Tabs.}
\newcommand{\system}{ChatVideo\xspace}
\newcommand\blfootnote[1]{
  \begingroup
  \renewcommand\thefootnote{}\footnote{#1}
  \addtocounter{footnote}{-1}
  \endgroup
}

\usepackage{etoolbox}
\makeatletter
\AfterEndEnvironment{algorithm}{\let\@algcomment\relax}
\AtEndEnvironment{algorithm}{\kern2pt\hrule\relax\vskip3pt\@algcomment}
\let\@algcomment\relax
\newcommand\algcomment[1]{\def\@algcomment{\footnotesize#1}}
\renewcommand\fs@ruled{\def\@fs@cfont{\bfseries}\let\@fs@capt\floatc@ruled
  \def\@fs@pre{\hrule height.8pt depth0pt \kern2pt}%
  \def\@fs@post{}%
  \def\@fs@mid{\kern2pt\hrule\kern2pt}%
  \let\@fs@iftopcapt\iftrue}
\makeatother
\begin{document}

\title{\system: A Tracklet-centric Multimodal and Versatile Video \\ Understanding System}

\author{Junke Wang$^{1,2}$,~Dongdong Chen$^{3}$, ~Chong Luo$^{4}$,~Xiyang Dai$^{3}$, \\ ~Lu Yuan$^{3}$, ~Zuxuan Wu$^{1,2\dagger}$, ~Yu-Gang Jiang$^{1,2\dagger}$
\\
$^{1}$Shanghai Key Lab of Intell. Info. Processing, School of CS, Fudan University \\
$^{2}$Shanghai Collaborative Innovation Center on Intelligent Visual Computing \\
$^{3}$Microsoft Cloud + AI, $^{4}$Microsoft Research Asia 
}

\maketitle

\begin{abstract}
\blfootnote{$^{\dagger}$ Corresponding authors.}
Existing deep video models are limited by specific tasks, fixed input-output spaces, and poor generalization capabilities, making it difficult to deploy them in real-world scenarios. In this paper, we present our vision for multimodal and versatile video understanding and propose a prototype system, \system. Our system is built upon a tracklet-centric paradigm, which treats tracklets as the basic video unit and employs various Video Foundation Models (ViFMs) to annotate their properties \eg, appearance, motion, \etc. All the detected tracklets are stored in a database and interact with the user through a database manager. We have conducted extensive case studies on different types of in-the-wild videos, which demonstrates the effectiveness of our method in answering various video-related problems. Our project is available at \href{https://www.wangjunke.info/ChatVideo/}{https://www.wangjunke.info/ChatVideo/}
\end{abstract}

\section{Introduction}
\label{sec:intro}
With the rise of various streaming platforms, videos are becoming a dominant component of Internet traffic, which has motivated the  development of deep learning-based video understanding techniques~\cite{huang2018makes,wang2022bevt,wang2023masked,wang2023look,wang2022omnivl,wang2023omnitracker,zhao2023streaming}. As one of the most important areas in computer vision, video understanding refers to the automatic extraction and interpretation of meaningful semantics from videos, which enjoys a wide range of applications such as online advertising and augmented reality.

Recently, video foundation models (ViFM)~\cite{bommasani2021opportunities,wang2022omnivl,wang2022internvideo,li2023unmasked} are gaining more and more attention in the community due to their superior performance on different video benchmarks and the pioneering exploration of unified video architectures. However, existing ViFMs still focus on a specific field, and none of them is capable of unifying all the video tasks. For example, OmniVL~\cite{wang2022omnivl} supports multimodal understanding and generation tasks, \eg, text-video retrieval and video captioning, while OmniTracker~\cite{wang2023omnitracker} addresses various tracking tasks like single object tracking (SOT) and multiple object tracking (MOT) with a fully shared network. The potential reason behind this lies in different types of video tasks relying on diversified feature modeling patterns and output heads, making the development of a One-For-All video model not only poses remarkable challenges but also consumes significant annotation and training costs. 

The emergence of large language models (LLMs)~\cite{brown2020language,chung2022scaling,zhang2022opt,touvron2023llama} represented by ChatGPT, however, provides a novel perspective for the solution to this problem. Visual ChatGPT~\cite{wu2023visual} is a pioneering work that makes use of the comprehension and reasoning capabilities of LLMs for the disentanglement of visual-related questions. Specifically, they propose to connect various Image Foundation Models (IFMs) to ChatGPT through a Prompt Manager, which decomposes the user question into a chain of instructions and then schedules different IFMs to answer it. This approach can be seen as a top-down visual understanding pattern, which first breaks down a composite task into several sub-tasks and then invokes the corresponding functions to resolve it step by step.

\begin{table*}[t]
\centering
  \begin{tabular*}{\linewidth}{@{\extracolsep{\fill}}lc | cc | c }
    \toprule
    \textbf{Type} && \textbf{Tasks} && \textbf{ViFMs} \\
    \midrule
    \multirow{3}{*}{Clip-based} && Action Recognition && \multirow{3}{*}{OmniVL~\cite{wang2022omnivl}, InternVideo~\cite{wang2022internvideo}, VATT~\cite{akbari2021vatt} \etc} \\
    ~ && (Dense) Video Captioning && ~ \\
    ~ && Temporal Action Localization && ~ \\ 
    \midrule
    \multirow{5}{*}{Instance-based} && Single Object Tracking && \multirow{5}{*}{Unicorn~\cite{yan2022towards}, OmniTracker~\cite{wang2023omnitracker}, UNINEXT~\cite{yan2023universal}, \etc} \\
    ~ && Video Object Segmentation && ~ \\
    ~ && Multiple Object Tracking (and Segmentation) && ~ \\
    ~ && Video Instance Segmentation && ~ \\
    ~ && Referred Video Object Segmentation && ~ \\
    \midrule
    \midrule
    \multirow{4}{*}{Audio Models} && Audio Classification && \multirow{4}{*}{\makecell[c]{CLAP~\cite{elizalde2022clap}, Hubert~\cite{hsu2021hubert}, Speech2Text~\cite{wang2020fairseq}, \\ UniSpeech~\cite{wang2021unispeech}, Wav2Vec~\cite{baevski2020wav2vec}, Whisper~\cite{radford2022robust}, \etc}} \\
    ~ && Automatic Speech Recognition && ~ \\
    ~ && Punctuation Restoration && ~ \\
    ~ && Speech Emotion Classification && ~ \\
    \bottomrule
  \end{tabular*}
  \vspace{-0.1in}
 \caption{Existing Video Foundation Models (ViFMs) and powerful audio models. }
\label{tab:video_tasks}
%\vspace{-0.1in}
\end{table*}

Extending the idea of Visual ChatGPT to the video domain, while promising, requires non-trivial effort due to the wealth of information in video data. Unlike images, videos record the changes in scenes over time, the movement of different objects, and the interaction between them. Therefore, the rich amount of information makes it rather inefficient to apply a single or several sequentially executed ViFMs for each user question. On the contrary, we argue a bottom-up pattern is more feasible for building a versatile video understanding system, \ie, comprehensively parsing the video first and adaptively querying the useful information during the interaction with users. To this end, we propose a trajectory-centric video understanding paradigm, where the tracklets of different object instances are treated as the basic unit of videos\footnote{In this paper, we use ``tracklet'' to refer to the ``instance'' it contains.}, and their attributes, \eg, appearance and trajectory, are predicted by different ViFMs. With the comprehensive annotations of different tracklets, the high-level semantics of a video could be easily obtained to answer various kinds of questions by the user.

Equipped with the proposed paradigm, we further present \system: a multimodal and versatile video understanding prototype system that enables a chat-based experience. We store the tracklets, as well as their categories, appearance, motion, and trajectories, in a database, and introduce a \textit{database manager} to translate user questions into standard database query commands. Finally, LLMs process, summarize, and polish the query results to provide the final neural language responses to the user. In this way, \system could communicate with the user in a conversational manner and offer relevant answers based on the context and specific problem at hand. The contributions of this work are summarized as follows:

\begin{itemize}
    \item We introduce \system, which combines abundant functions of various ViFMs, along with the conversational and reasoning capability of ChatGPT for multimodal and versatile video understanding.
    \item We propose a novel tracklet-centric paradigm, which interprets the video contents with the basic ``tracklet" element. The abundant attributes of different tracklets in a video are predicted by different ViFMs and then stored in a database, and a database manager serves as a bridge between the database and the users.
    \item Extensive case studies are conducted to evaluate the performance of our method and study its behavior, which demonstrates the effectiveness of our system in addressing various video-related questions and showcases its potential for real-world applications. 
\end{itemize}

\begin{figure*}[t]
  \centering
   \includegraphics[width=\linewidth]{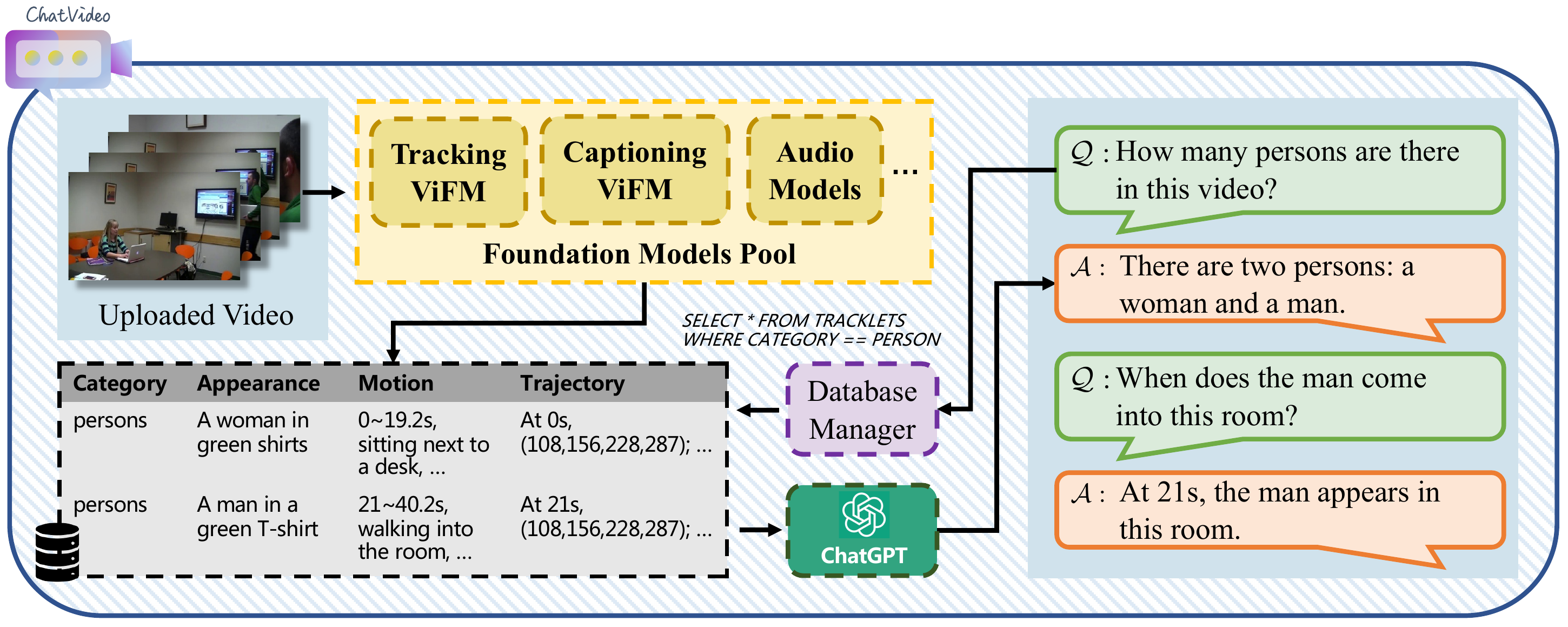}
   \vspace{-0.1in}
   \caption{Overview of \system. For an uploaded video, \system first detects all the tracklets in it and recognizes their category, appearance, motion, \etc, which are all stored in a database. After the user enters a question like ``how many persons are there in this video'', the proposed database manager converts it into a query statement, and retrieves the useful information from the above-mentioned database. Finally, ChatGPT summarizes the query results and polishes them to a natural language description. Note that the components marked with dotted lines are not visible to the users.}
   \label{fig:architecture}
   %\vspace{-0.1in}
\end{figure*}

\section{Related Work}
\label{sec:related}
\subsection{Video Foundation Models}
Depending on the type of downstream tasks of interest, existing Video Foundation Models (ViFMs) could be classified into clip-based ViFMs and instance-based ViFMs. The former~\cite{wang2022omnivl,wang2022internvideo,wang2022allinone,akbari2021vatt} is skilled in summarizing the content of videos for sequence-level decisions like action recognition, text-to-video retrieval, and video captioning, while the latter~\cite{yan2022towards,yan2023universal,wang2023omnitracker}, which are dedicated to the instance understanding for dense prediction tasks, \ie, visual object tracking. We summarize different types of ViFMs in Table~\ref{tab:video_tasks}. Considering audio is a key piece of information in understanding the content of the video like the mood of the characters, we also list the powerful audio models~\cite{baevski2020wav2vec,wang2020fairseq,elizalde2022clap,hsu2021hubert,ao2021speecht5,wang2021unispeech,babu2021xls,chen2022unispeech,chen2022wavlm,radford2022robust}. Although more and more downstream tasks can be accommodated into one model, the differentiated feature modeling approaches and output space still hinder the development of a truly unified foundation model for all video tasks. In this work, we try to implement a universal video understanding system by integrating different ViFMs and audio models to fully utilize their expertise in a particular area.

\subsection{Interactive Video Understanding Systems}
An ideal interactive video understanding system should be able to chat with the end users based on the video content. The most relevant topic that has been widely explored in academia is video question answering (Video QA)~\cite{lei2018tvqa}, which aims to answer natural language questions according to the given video. However, deploying existing Video QA models for interactive video understanding systems faces twofold challenges: 1) limited by the annotations of mainstream benchmarks, Video QA methods can only answer individual questions and lack the ability to relate the conversational contexts, making it difficult for users to enjoy a communicative experience. 2) since most Video QA approaches~\cite{lei2019tvqa+,wang2022omnivl,xu2023mplug,chen2023valor} only reply on global video features to represent a video, they may struggle with the complex questions that require fine-grained understanding ability. The above problems motivate us to present our vision for an interactive video understanding system in this work, \ie, combining various deep video models with ChatGPT, so as to leverage its conversational capabilities to enable interaction with users.

\subsection{Connecting ChatGPT to Visual Models}
Since the emergence of ChatGPT, numerous researchers from the computer vision field have been exploring ways to integrate it with existing deep visual models to implement innovative applications. Visual ChatGPT~\cite{wu2023visual} pioneers the idea of bridging ChatGPT and various image foundation models (IFMs) \cite{rombach2022high,radford2021learning,yuan2021florence} through a prompt manager, which schedules IFMs with different functions according to the user inputs. Based on this, TaskMatrix.AI~\cite{liang2023taskmatrix} and HuggingGPT~\cite{shen2023hugginggpt} further illustrate their vision for building a novel AI ecosystem that connects ChatGPT to millions of APIs or AI models available in Hugging Face that own diversified textual and visual processing skills. In addition, Video ChatCaptioner~\cite{chen2023video} also presents the possibility to annotate video data, where ChatGPT functions as a questioner and BLIP2~\cite{li2023blip} functions as an answerer. In this paper, we present the first attempt at applying ChatGPT to interactive video understanding. Different ViFMs cooperate with each other to detect the tracklets and annotate their attributes in a video, while ChatGPT is responsible for information retrieval, processing, reasoning, and interaction with users.

\section{Method}
\label{sec:method}
Our goal is to build a multimodal and versatile video understanding system that offers comprehensive functionality and excellent interactivity. To this end, we propose to interpret the video contents with a novel tracklet-centric paradigm and present a prototype system, \system. In this section, we first briefly review ChatGPT and the ViFMs that we use in Sec.~\ref{subsec:pre}, and then introduce the pipeline of our system in Sec.~\ref{subsec:arch}. Finally, we illustrate how \system interacts with users in Sec.~\ref{subsec:interact}. Figure~\ref{fig:architecture} gives an overview of the framework of \system.

\subsection{Preliminaries}
\label{subsec:pre}
\system maintains a \textbf{Foundation Models Pool} to store various video foundation models, which own the capability to detect and annotate different attributes of tracklets in videos. Here we list several models in it:

\vspace{0.05in}
\noindent \textbf{OmniTracker}~\cite{wang2023omnitracker} is an instance-based video foundation model, which addresses five tracking tasks, \ie, Single Object Tracking (SOT), Video Object Segmentation (VOS), Multiple Object Tracking (MOT), Multiple Object Tracking and Segmentation (MOTS), and Video Instance Segmentation (VIS) with a fully shared network architecture, model weights, and
inference pipeline. Specifically, They propose a tracking-with-detection paradigm, where tracking supplements appearance priors for the detection in the current frame, and detection provides tracking with candidate bounding boxes for the temporal association. 

\vspace{0.05in}
\noindent \textbf{OmniVL}~\cite{wang2022omnivl} is a clip-based video foundation model based on vision-language pretraining. It follows an encoder-decoder structure, where two unimodal encoders are adopted to extract the visual and text representations, a visual-grounded alignment decoder for semantic alignment discrimination, and a visual-grounded generation decoder for open-ended text generation, respectively. Notably, OmniVL is the first work that supports both image (frame) tasks and video (clip) tasks in a single model, making it an ideal candidate for building a universal video understanding system as appearance and motion annotations are equally important to comprehensively and accurately describe the video contents.

\vspace{0.05in}
\noindent \textbf{Whisper}~\cite{radford2022robust} is an automatic speech recognition system, which learns robust speech representations from large-scale multilingual data crawled from the Internet.

\vspace{0.05in}
\noindent \textbf{Wav2Vec 2.0}~\cite{baevski2020wav2vec} learns strong audio representations from raw audio data in a self-supervised manner. It could be fine-tuned to solve various audio-related tasks like audio classification and emotion classification.

It is worth noting that \system is designed to be extensible and more video foundation models can be integrated into it in the future to support more functions with superior performance.
 
\begin{figure*}[t]
  \centering
   \includegraphics[width=\linewidth]{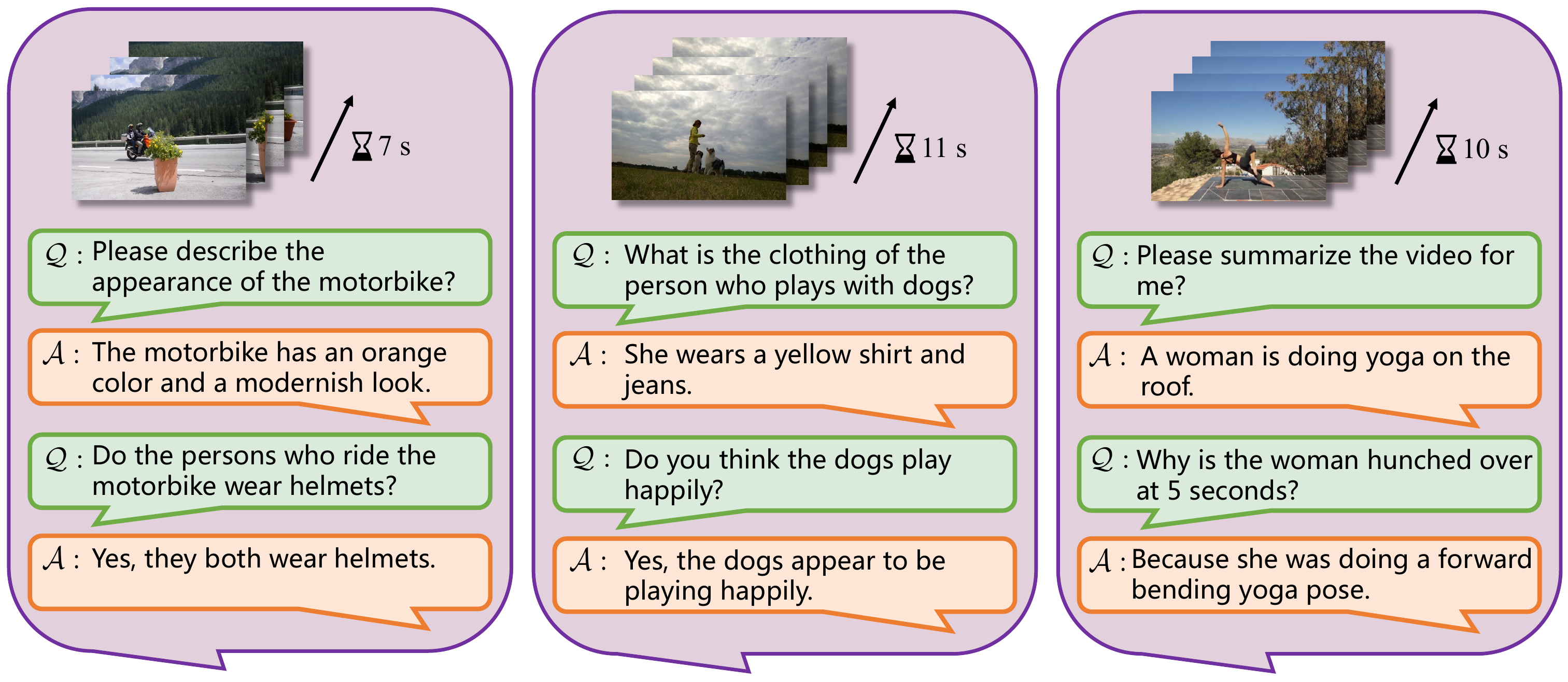}
   \vspace{-0.15in}
   \caption{\system for appearance-related questions.}
   \label{fig:case_appearance}
\end{figure*}

\subsection{Pipeline of \system}
\label{subsec:arch}

\noindent \textbf{Overview}. Given an uploaded video $\mathcal{V}$ of $s$ seconds, we first extract the audio $A$ and a sequence of frames $[\textit{X}_{1}, \textit{X}_{2}, ..., \textit{X}_{T}]$ from it with open-source tools like $\mathrm{ffmpeg}$~\cite{ffmpeg}, where $T$ denotes the number of frames and $T = \rm{FPS} \times s$. Then we detect the tracklets in $\mathcal{V}$ and predict their attributes, \eg, appearance, motion, and trajectories with the ViFMs mentioned above. Finally, we store all the tracklets in a database, and retrieve the relevant information from the database during the interaction with users through a proposed database manager.

\vspace{0.05in}
\noindent \textbf{Tracklet Database Construction}. Taking the video frames $[\textit{X}_{1}, \textit{X}_{2}, ..., \textit{X}_{T}]$ as input, OmniTracker first detects the tracklets $\{ K_{i} = (c_{i}, \{b_{j})\}_{j=st_{i}}^{ed_{i}})\}_{i=1}^{N}$ in it, where $N$ is the number of tracklets detected, $c_{i}$ denotes the category of the $i$-th tracklet, $\{b_{j})\}_{j=st_{i}}^{ed_{i}}$ is the bounding box in $st_{i}$-th $\sim$ $ed_{i}$-th frame. Then we crop the $i$-th tracklet from the video frames to form a spatial-temporal tracklet, \ie, $R_{i} = \{(t_{j}, r_{j})\}_{j=st_{i}}^{ed_{i}}$, where $t_{j}$ is the timestamp measured in s that can be calculated by $t_{j} = j / \rm{FPS}$, $r_{j}$ is the region cropped from the $j$-th frame. Note that we additionally append the entire video as a special tracklet, representing the environment in which the video takes place.

After that, we employ OmniVL to predict the attributes of $i$-th tracklet through image and video captioning. For the \textit{appearance} information, we caption the region in $R_{i}$ at different time steps with the prompt ``What does the $\{cat_{i}\}$ look like? The $\{cat_{i}\}$''. While for the \textit{temporal dynamics}, we split $R_{i}$ into several segments with equal length, and then utilize OmniVL to caption them with the prompt "What is the $\{cat_{i}\}$ doing? The $\{cat_{i}\}$". We also classify the audio $A$ with an audio classification model~\cite{mit} finetuned on AudioSet~\cite{gemmeke2017audio}, if the category is ``speech'' related, we further apply multi-lingual ASR model Whisper \cite{radford2022robust} to recognize the speech contents and finetuned Wav2Vec2 \cite{baevski2020wav2vec} model~\cite{superb} to predict the emotion of the speaker. With this, we build a database $\mathcal{D}$ to store the fine-grained information of all the detected tracklets. The field names and values in $\mathcal{D}$ are defined in the following format:
\begin{itemize}[leftmargin=*]
\setlength{\itemsep}{1.5pt}
\setlength{\parsep}{0pt}
\setlength{\parskip}{0pt}
    \item $\mathrm{ID}$: the primary key that uniquely identifies a record.
    \item $\mathrm{Category}$: the category of the $i$-th tracklet (instance), predicted by OmniTracker.
    \item $\mathrm{Appearance}$: the appearance of the $i$-th tracklet (instance), predicted by OmniVL with image captioning.
    \item $\mathrm{Motion}$: the motion of the $i$-th tracklet (instance) in different temporal segments. We append the timestamp before each motion description, and the format is ``from \{segment start\} to \{segment end\} s, the \{category\} is \{description of the motion\}''.
    \item $\mathrm{Trajectory}$: the trajectory of the $i$-th tracklet (instance), predicted by Omnitracker. The timestamp is also added: ``at \{time of the current frame\}, the \{category of the instance\} locates at \{coordinate\}.
    \item $\mathrm{Audio}$: the category of the audio in the given video, the content and emotion of the speakers are optionally included. This field is only available for the tracklet which contains the complete video.
\end{itemize}
We show several examples of the database in Sec.~\ref{sec:exp}.

\subsection{Interaction with Users}
\label{subsec:interact}
As mentioned above, the conversational and reasoning capability of ChatGPT equips our system with the ability to interact with users. Following Visual ChatGPT~\cite{wu2023visual}, we first define the principle for $\rm{ChatGPT}$ \footnote{In this paper, we use $\rm{ChatGPT}$ to refer to an initialized ChatGPT during the conversation with a specific user.} to constrain its code of conduct and reasoning process, \eg, it should be sensitive to the path of a video. Next, during the conversation with a user, let $Q_{i}$ denote the $i$-th question from the user, we employ a database manager $\mathcal{M}$ to convert it to a database query $\widetilde{Q}_{i}$: $\widetilde{Q}_{i} = \mathcal{M}(Q_{i})$. The database manager is also driven by a large language model (LLM) and expert in translating the user question into a proper query command. Note that the past conversations are not input to the database manager so that it will return as many relevant records as possible, and the selection of the retrieved results and the association with the context is completed by $\rm{ChatGPT}$. Then $\widetilde{Q}_{i}$ could be used to retrieve the relevant information $S_{i}$ from $\mathcal{D}$.

In the following, we input $Q_{i}$, $S_{i}$, the dialogue history $H_{i} = \{(Q_{j}, A_{j})\}_{j=1}^{i-1}$ to $\rm{ChatGPT}$, which returns us the answer $A_{i}$, which is described in natural language. As above-mentioned, more foundation models other than OmniVL, OmniTracker, and Whisper, could be added to the foundation model pool. With more models involved, when $\rm{ChatGPT}$ finds that the results retrieved from the constructed $\mathcal{D}$ are empty or insufficient to give an accurate answer, it can work like the prompt manager in Visual ChatGPT~\cite{wu2023visual} and employ other models in the Foundation Model Pool to get more information.

\begin{figure*}[t]
  \centering
   \includegraphics[width=\linewidth]{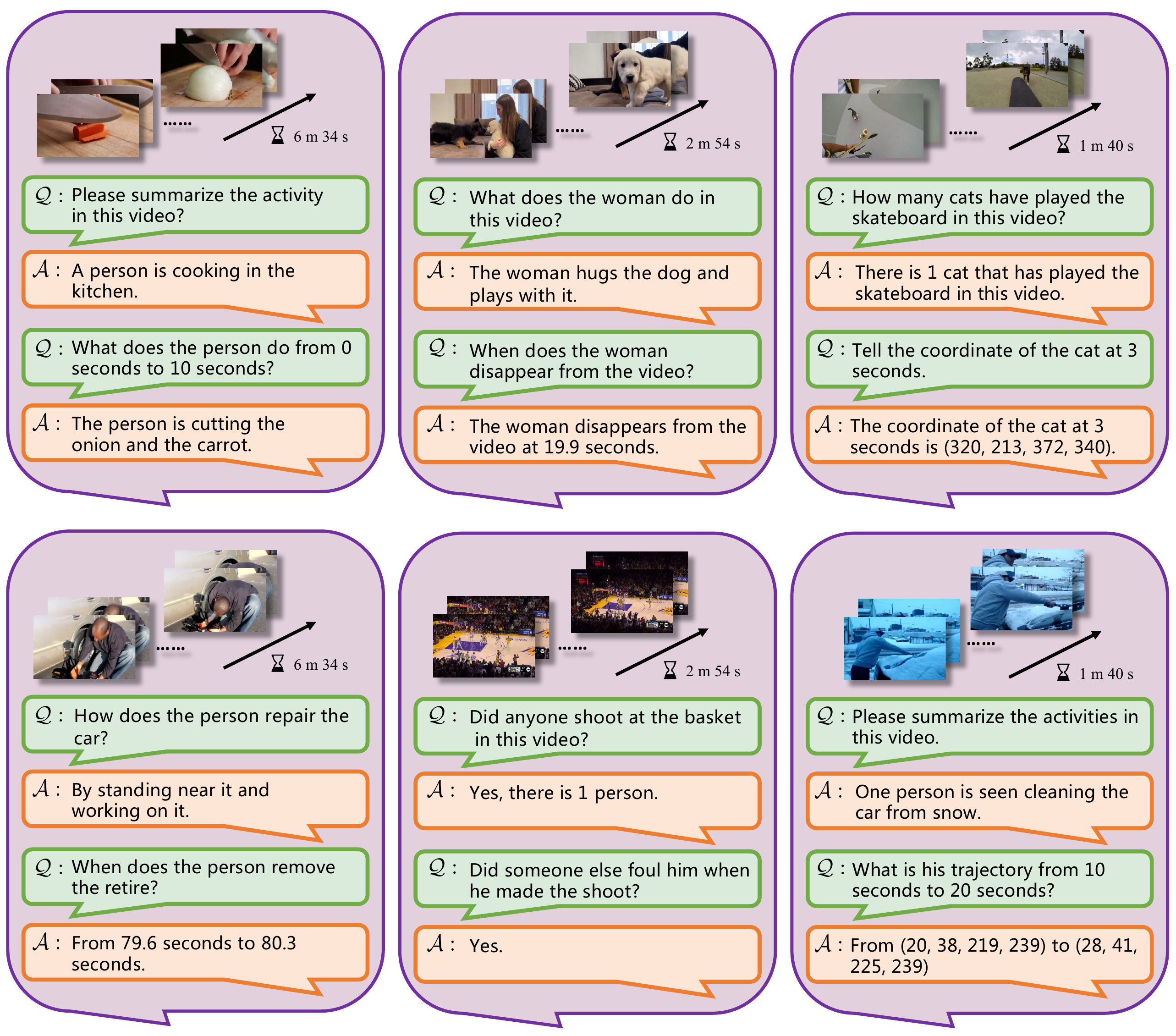}
    \vspace{-0.1in}
   \caption{\system for motion-related questions.}
   \label{fig:case_motion}
\end{figure*}

\section{Experiments}
\label{sec:exp}

\noindent \textbf{Implementations.} In order to improve the response efficiency of our system, we select only one frame in each tracklet to annotate its appearance. The selection strategy is based on both the size of the bounding box and its distance to the boundary in each frame, \ie, $\mathop{\rm{argmax}}\limits_{j} (\sqrt{\rm{area}(b_{j})} + \rm{dist}(b_{j}))$. While for the motion of different tracklets, the length of the temporal segments is 32 frames. We adopt the pretrained OmniVL~\cite{wang2022omnivl} for video captioning, and fine-tuned model on COCO~\cite{lin2014microsoft} for image captioning. For the audio models, we adopt  open-sourced models for audio classification~\cite{mit}, automatic speech recognition~\cite{radford2022robust}, and emotion classification~\cite{superb}.

\begin{figure*}[t]
  \centering
   \includegraphics[width=\linewidth]{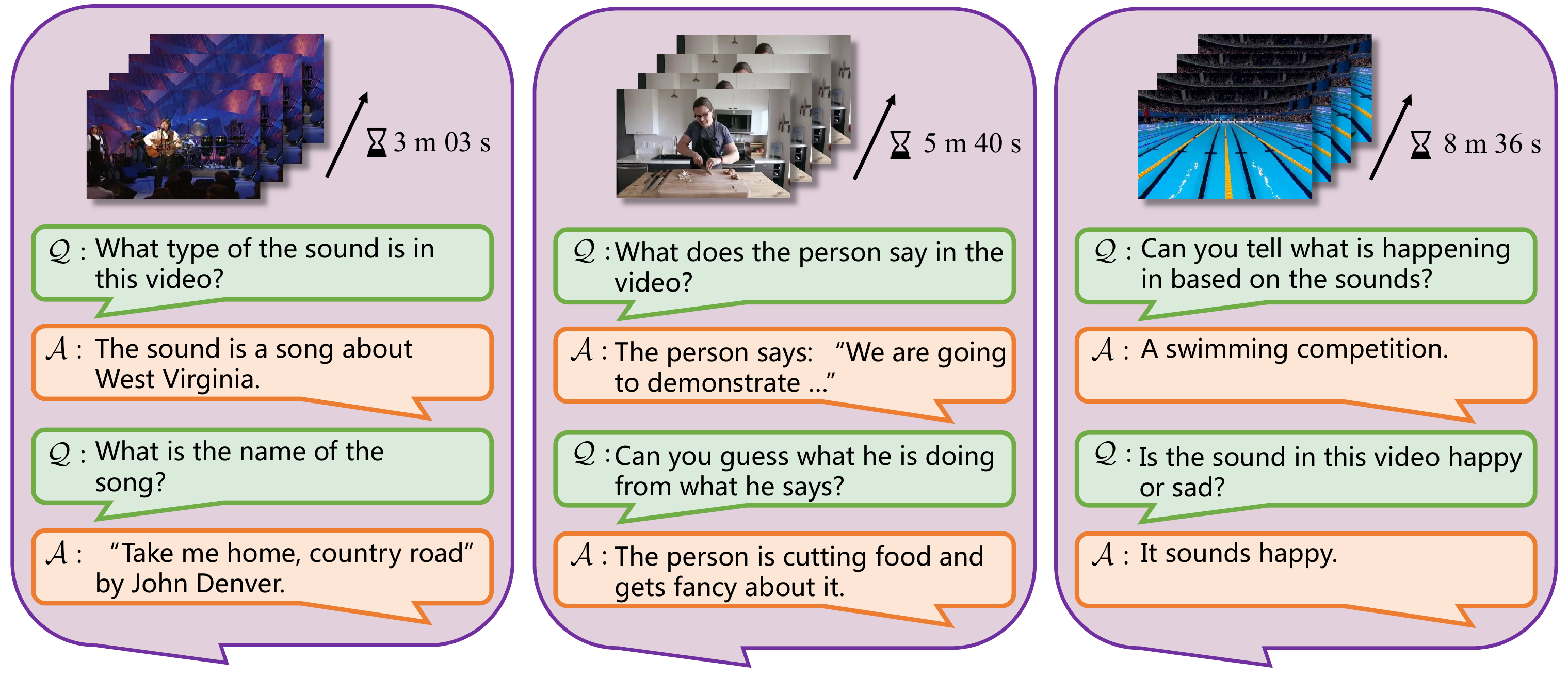}
    \vspace{-0.15in}
   \caption{\system for audio-related questions.}
   \label{fig:case_audio}
\end{figure*}

\subsection{Case Study for Appearance-related Questions}
We first evaluate the performance of \system in answering appearance-related questions, and the results are shown in Figure~\ref{fig:case_appearance}. We can see that our system could provide detailed information about objects and people within a video, including their color, motivation, and even actions at particular moments. These capabilities demonstrate the versatility and effectiveness of our system for a variety of applications such as content analysis, safety checks, and recommendations.

\subsection{Case Study for Motion-related Questions}
The temporal modeling ability of \system is then evaluated by asking it motion-related questions on long videos. In contrast to short video clips, long videos typically have more complex content and require more memory consumption, making them more challenging for video understanding systems to analyze. However, our system, which is built on several powerful ViFMs, could process the videos in an online manner, and understand the semantic information from bottom to up.

The results in Figure~\ref{fig:case_motion} show that \system could accurately summarize the activities in a video, recognize the events within a specific time period, and predict the locations (and trajectories) of different objects. These showcase the potential of our system for crowd counting, long video captioning, and (anomaly) event detection.

\subsection{Case Study for Audio-related Questions}
Audio is also a piece of important information in videos, which records what the speaker says and even reflects his/her emotion. As shown in Figure~\ref{fig:case_audio}, \system could recognize the type of audio and its contents, predict the emotion of the sounds, \etc.  

\subsection{Visualization of Tracking Results and Database}
\system is built upon the tracklet-centric video understanding paradigm, therefore, the performance of tracking models may have a prominent influence on the answers \system feeds back to the users. We visualize the tracking results predicted by OmniTracker, as well as the annotations produced by OmniVL in Table~\ref{fig:vis}. {Note that we omit the field  of audio for brevity.}. The results demonstrate \system could track various types of instances in the videos and generate accurate captions for their fine-grained attributes like appearance and motion. 

\subsection{System Principle for the Dataset Manager}
As mentioned in Sec.~\ref{subsec:interact}, the database manager is driven by an LLM. We define the following prompt for it:
\small
\label{alg:sample}
\algcomment{\fontsize{7.2pt}{0em}\selectfont \texttt{linespace}: generate evenly spaced values
}
\definecolor{codeblue}{rgb}{0.25,0.5,0.5}
\definecolor{codegreen}{rgb}{0,0.6,0}
\definecolor{codekw}{rgb}{0.85, 0.18, 0.50}
\lstset{
  backgroundcolor=\color{white},
  basicstyle=\fontsize{7.5pt}{7.5pt}\ttfamily\selectfont,
  columns=fullflexible,
  breaklines=true,
  captionpos=b,
  commentstyle=\fontsize{7.5pt}{7.5pt}\color{codekw},
  keywordstyle=\fontsize{7.5pt}{7.5pt}\color{codekw},
  escapechar={|}, 
}
\begin{lstlisting}[language=python]
"""Given an input question, first create a syntactically correct {dialect} query to run, then look at the results of the query and return the answer. Use the following format:

Question: "Question here"
SQLQuery: "SQL Query to run"
SQLResult: "Result of the SQLQuery"
Answer: "Final answer here"

Only use the following tables:
{table_info}

The records in the tables are in the following format:

```
ID: the primary key of the record.
Category: the category of the tracklet.
Appearance: the appearance of the tracklet.
Motion: the motion of the tracklet, described as "from t1 to t2 seconds, movements of the object".
Trajectory: the trajectory of the tracklet, described as "at t seconds, (x1, y1, x2, y2)". The velocity of the object could be obtained by calculating the distance between two positions.
Audio: the audio in this video
```

The records in the tables are randomly ordered. If the results of the SQLQuery include multiple records, you should list them separately in your answers instead of mixing them together.

Question: {input}"""
\end{lstlisting}

\begin{figure*}[t]
\centering
\includegraphics[width=\linewidth]{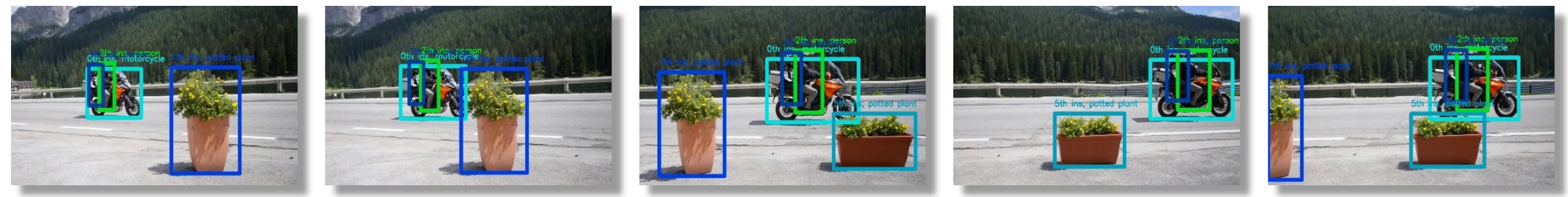}
\vspace{-0.2in}
\end{figure*}

\begin{table*}[!ht]
\centering
  \begin{tabular*}{\linewidth}{|p{0.1in} | c | p{1in} | p{2.5in} | p{1.7in}|}
    \toprule
    \textbf{ID} & \textbf{Category} & \textbf{Appearance} & \textbf{Motion} & \textbf{Trajectory} \\
    \midrule
     0 & environment & road and mountains & From 0 to 7 s, a motorcyclist riding on the road in the mountains; ... &  N/A \\
    \midrule
    1 & motorcycle &  orange in color &  From 0 to 7 s, a man riding a motorcycle down a road; ... &  at 0 s, (198,198,294,277); ... \\
    \midrule
    2 & person & wearing a black leather jacket and a black helmet & From 0 to 7 s, the person is a motorcyclist on a motorcycle in the mountains; ... & at 0 s, (222,176,279,259); ... \\
    \midrule
    ... & ... & ... & ... & ... \\
    \bottomrule
  \end{tabular*}
\end{table*}

\begin{figure*}[!ht]
\centering
\includegraphics[width=\linewidth]{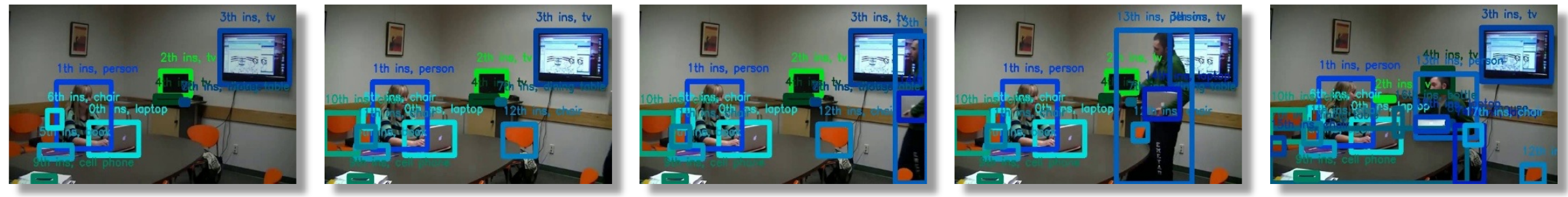}
\vspace{-0.25in}
\end{figure*}

\begin{table*}[!ht]
\centering
  \begin{tabular*}{\linewidth}{|p{0.1in} | c | p{1in} | p{2.5in} | p{1.7in}|}
    \toprule
    \textbf{ID} & \textbf{Category} & \textbf{Appearance} & \textbf{Motion} & \textbf{Trajectory} \\
    \midrule
     0 & environment &  a classroom &  From 0 to 1.1 s, a woman is sitting in the room; ... & N/A \\
    \midrule
     1 & laptop &  laptop black and silver in color &  From 0 to 1.1 s, a person is working on a laptop; ... &  at 0 s, (181,236,289,300); ... \\
    \midrule
    2 & person & person long hair and green T-shirt & From 0 to 1.1 s, the person is a woman in the classroom; ... & at 0 s, (122,159,225,289); ... \\
    \midrule
    3 & tv & tv black screen & From 0 to 1.2 s, the tv is on a black background; ... & at 0 s, (338,133,406,181); ... \\
    \midrule
    ... & ... & ... & ... & ... \\
    \bottomrule
   \end{tabular*}
   
   \caption{Visualization of the tracking results and the constructed dataset on two videos.}
   \label{fig:vis}
\end{table*}

\subsection{Failure Cases and Analysis}
\noindent \textbf{Tracking for Fast-moving Objects:} Tracking for fast-moving objects is a remarkable challenge for the tracking community. Since existing tracking models~\cite{wang2023omnitracker,yan2022towards,yan2023universal} are vulnerable to such cases, our system may give inaccurate answers to the counting questions, or the questions that depend heavily on temporal correlation.

\vspace{0.05in}
\noindent \textbf{Fine-grained Action Recognition:} While our system is effective at recognizing broad categories of actions, such as walking or jumping, it may struggle with fine-grained action recognition, such as distinguishing between different types of jumps or identifying more subtle movements.

\vspace{0.05in}
\noindent \textbf{Audio Classification:} Limited by the generalization capability of the audio classification models, our system may fail to predict the category of sounds in videos. Therefore, in our implementation, we relaxed the condition that whatever the category of the audio is, we will use the ASR model~\cite{radford2022robust} to identify the content in it.

\section{Conclusion and Future Work}
\label{sec:conclusion}
This paper presents \system, a novel video understanding system that combines the capabilities of ChatGPT and ViFMs to achieve multimodal and versatile video understanding. Our approach is based on a tracklet-centric paradigm, where the tracklet is considered the basic unit for analyzing the video content, and its properties, \eg, appearance and motion, are predicted by different ViFMs. We store the parsed information of different tracklets in a database, so as to utilize it in a more flexible manner. During the interaction with users, a database manager functions as a bridge to translate their questions into database queries, the results of which are summarized and polished by ChatGPT to obtain a natural language description. Through extensive case studies, we demonstrate the effectiveness of our approach in addressing various video-related questions and scenarios, which validate the potential of our system for real-world applications such as video content recommendation and online education.

Our work represents a significant step towards the development of multimodal and versatile video understanding systems, and we anticipate that it will motivate further research in this area. We identify several open problems that merit exploration, along with potential future directions:

\vspace{0.05in}
\noindent \textbf{More powerful video foundation models:} We observe that the existing ViFMs oftentimes struggle on in-the-wild videos where the scenes might be rather complicated and the camera viewpoints might have significant changes, revealing their poor generalization ability. To address this, future research directions could include jointly training ViFMs on a broader range of tasks, or collecting more higher-quality data.

\vspace{0.05in}
\noindent \textbf{Robustness against adversarial attacks:} A critical challenge for the deployment of video understanding systems is to guarantee their robustness against adversarial attacks. One possible direction for future work is to investigate the use of adversarial training or robust optimization techniques to improve the robustness of video understanding systems. Another direction is to develop new methods for detecting and mitigating adversarial attacks in video data. For example, one could explore the use of explainable AI techniques to better understand the behavior of video understanding systems and detect when they are being manipulated by adversarial attacks.

\vspace{0.05in}
\noindent \textbf{Efficiency:} Efficiency is a crucial factor for any practical video understanding system~\cite{wang2022efficient}, where the satisfying interactive experience is built on a fast response. One possible direction is to explore the design of efficient video models, such as lightweight Transformer models or network pruning. Another direction is to investigate the use of hardware acceleration, such as GPUs or specialized hardware, to speed up the inference process.

\vspace{0.05in}
\noindent \textbf{Training with RLHF mechanism:} Another promising direction is to investigate the use of reinforcement learning with human feedback (RLHF) mechanisms to train video understanding systems. RLHF has been shown to be effective in improving the performance of natural language processing systems, and it could be similarly effective in training video understanding systems. For example, one could use RLHF to optimize the response generation process in Video ChatGPT or to optimize the tracklet selection process in the tracklet-centric paradigm.

\clearpage

{\small
\bibliographystyle{ieee_fullname}
\bibliography{egbib}

\begin{thebibliography}{10}\itemsep=-1pt

\bibitem{akbari2021vatt}
Hassan Akbari, Liangzhe Yuan, Rui Qian, Wei-Hong Chuang, Shih-Fu Chang, Yin
  Cui, and Boqing Gong.
\newblock Vatt: Transformers for multimodal self-supervised learning from raw
  video, audio and text.
\newblock In {\em NeurIPS}, 2021.

\bibitem{ao2021speecht5}
Junyi Ao, Rui Wang, Long Zhou, Chengyi Wang, Shuo Ren, Yu Wu, Shujie Liu, Tom
  Ko, Qing Li, Yu Zhang, et~al.
\newblock Speecht5: Unified-modal encoder-decoder pre-training for spoken
  language processing.
\newblock {\em arXiv preprint arXiv:2110.07205}, 2021.

\bibitem{babu2021xls}
Arun Babu, Changhan Wang, Andros Tjandra, Kushal Lakhotia, Qiantong Xu, Naman
  Goyal, Kritika Singh, Patrick von Platen, Yatharth Saraf, Juan Pino, et~al.
\newblock Xls-r: Self-supervised cross-lingual speech representation learning
  at scale.
\newblock {\em arXiv preprint arXiv:2111.09296}, 2021.

\bibitem{baevski2020wav2vec}
Alexei Baevski, Yuhao Zhou, Abdelrahman Mohamed, and Michael Auli.
\newblock wav2vec 2.0: A framework for self-supervised learning of speech
  representations.
\newblock In {\em NeurIPS}, 2020.

\bibitem{bommasani2021opportunities}
Rishi Bommasani, Drew~A Hudson, Ehsan Adeli, Russ Altman, Simran Arora, Sydney
  von Arx, Michael~S Bernstein, Jeannette Bohg, Antoine Bosselut, Emma
  Brunskill, et~al.
\newblock On the opportunities and risks of foundation models.
\newblock {\em arXiv preprint arXiv:2108.07258}, 2021.

\bibitem{brown2020language}
Tom Brown, Benjamin Mann, Nick Ryder, Melanie Subbiah, Jared~D Kaplan, Prafulla
  Dhariwal, Arvind Neelakantan, Pranav Shyam, Girish Sastry, Amanda Askell,
  et~al.
\newblock Language models are few-shot learners.
\newblock In {\em NeurIPS}, 2020.

\bibitem{chen2023video}
Jun Chen, Deyao Zhu, Kilichbek Haydarov, Xiang Li, and Mohamed Elhoseiny.
\newblock Video chatcaptioner: Towards the enriched spatiotemporal
  descriptions.
\newblock {\em arXiv preprint arXiv:2304.04227}, 2023.

\bibitem{chen2023valor}
Sihan Chen, Xingjian He, Longteng Guo, Xinxin Zhu, Weining Wang, Jinhui Tang,
  and Jing Liu.
\newblock Valor: Vision-audio-language omni-perception pretraining model and
  dataset.
\newblock {\em arXiv preprint arXiv:2304.08345}, 2023.

\bibitem{chen2022wavlm}
Sanyuan Chen, Chengyi Wang, Zhengyang Chen, Yu Wu, Shujie Liu, Zhuo Chen, Jinyu
  Li, Naoyuki Kanda, Takuya Yoshioka, Xiong Xiao, et~al.
\newblock Wavlm: Large-scale self-supervised pre-training for full stack speech
  processing.
\newblock {\em STSP}, 2022.

\bibitem{chen2022unispeech}
Sanyuan Chen, Yu Wu, Chengyi Wang, Zhengyang Chen, Zhuo Chen, Shujie Liu, Jian
  Wu, Yao Qian, Furu Wei, Jinyu Li, et~al.
\newblock Unispeech-sat: Universal speech representation learning with speaker
  aware pre-training.
\newblock In {\em ICASSP}, 2022.

\bibitem{chung2022scaling}
Hyung~Won Chung, Le Hou, Shayne Longpre, Barret Zoph, Yi Tay, William Fedus,
  Eric Li, Xuezhi Wang, Mostafa Dehghani, Siddhartha Brahma, et~al.
\newblock Scaling instruction-finetuned language models.
\newblock {\em arXiv preprint arXiv:2210.11416}, 2022.

\bibitem{elizalde2022clap}
Benjamin Elizalde, Soham Deshmukh, Mahmoud~Al Ismail, and Huaming Wang.
\newblock Clap: Learning audio concepts from natural language supervision.
\newblock {\em arXiv preprint arXiv:2206.04769}, 2022.

\bibitem{ffmpeg}
ffmpeg.
\newblock A complete, cross-platform solution to record, convert and stream
  audio and video.
\newblock \url{https://ffmpeg.org}.

\bibitem{gemmeke2017audio}
Jort~F Gemmeke, Daniel~PW Ellis, Dylan Freedman, Aren Jansen, Wade Lawrence,
  R~Channing Moore, Manoj Plakal, and Marvin Ritter.
\newblock Audio set: An ontology and human-labeled dataset for audio events.
\newblock In {\em ICASSP}, 2017.

\bibitem{hsu2021hubert}
Wei-Ning Hsu, Benjamin Bolte, Yao-Hung~Hubert Tsai, Kushal Lakhotia, Ruslan
  Salakhutdinov, and Abdelrahman Mohamed.
\newblock Hubert: Self-supervised speech representation learning by masked
  prediction of hidden units.
\newblock {\em ASLP}, 2021.

\bibitem{huang2018makes}
De-An Huang, Vignesh Ramanathan, Dhruv Mahajan, Lorenzo Torresani, Manohar
  Paluri, Li Fei-Fei, and Juan~Carlos Niebles.
\newblock What makes a video a video: Analyzing temporal information in video
  understanding models and datasets.
\newblock In {\em CVPR}, 2018.

\bibitem{lei2018tvqa}
Jie Lei, Licheng Yu, Mohit Bansal, and Tamara~L Berg.
\newblock Tvqa: Localized, compositional video question answering.
\newblock {\em arXiv preprint arXiv:1809.01696}, 2018.

\bibitem{lei2019tvqa+}
Jie Lei, Licheng Yu, Tamara~L Berg, and Mohit Bansal.
\newblock Tvqa+: Spatio-temporal grounding for video question answering.
\newblock {\em arXiv preprint arXiv:1904.11574}, 2019.

\bibitem{li2023blip}
Junnan Li, Dongxu Li, Silvio Savarese, and Steven Hoi.
\newblock Blip-2: Bootstrapping language-image pre-training with frozen image
  encoders and large language models.
\newblock {\em arXiv preprint arXiv:2301.12597}, 2023.

\bibitem{li2023unmasked}
Kunchang Li, Yali Wang, Yizhuo Li, Yi Wang, Yinan He, Limin Wang, and Yu Qiao.
\newblock Unmasked teacher: Towards training-efficient video foundation models.
\newblock {\em arXiv preprint arXiv:2303.16058}, 2023.

\bibitem{liang2023taskmatrix}
Yaobo Liang, Chenfei Wu, Ting Song, Wenshan Wu, Yan Xia, Yu Liu, Yang Ou, Shuai
  Lu, Lei Ji, Shaoguang Mao, et~al.
\newblock Taskmatrix. ai: Completing tasks by connecting foundation models with
  millions of apis.
\newblock {\em arXiv preprint arXiv:2303.16434}, 2023.

\bibitem{lin2014microsoft}
Tsung-Yi Lin, Michael Maire, Serge Belongie, James Hays, Pietro Perona, Deva
  Ramanan, Piotr Doll{\'a}r, and C~Lawrence Zitnick.
\newblock Microsoft coco: Common objects in context.
\newblock In {\em ECCV}, 2014.

\bibitem{mit}
MIT.
\newblock Mit/ast-finetuned-audioset-10-10-0.4593.
\newblock \url{https://huggingface.co/MIT/ast-finetuned-audioset-10-10-0.4593.}

\bibitem{radford2021learning}
Alec Radford, Jong~Wook Kim, Chris Hallacy, Aditya Ramesh, Gabriel Goh,
  Sandhini Agarwal, Girish Sastry, Amanda Askell, Pamela Mishkin, Jack Clark,
  et~al.
\newblock Learning transferable visual models from natural language
  supervision.
\newblock In {\em International conference on machine learning}, pages
  8748--8763. PMLR, 2021.

\bibitem{radford2022robust}
Alec Radford, Jong~Wook Kim, Tao Xu, Greg Brockman, Christine McLeavey, and
  Ilya Sutskever.
\newblock Robust speech recognition via large-scale weak supervision.
\newblock {\em arXiv preprint arXiv:2212.04356}, 2022.

\bibitem{rombach2022high}
Robin Rombach, Andreas Blattmann, Dominik Lorenz, Patrick Esser, and Bj{\"o}rn
  Ommer.
\newblock High-resolution image synthesis with latent diffusion models.
\newblock In {\em Proceedings of the IEEE/CVF Conference on Computer Vision and
  Pattern Recognition}, pages 10684--10695, 2022.

\bibitem{shen2023hugginggpt}
Yongliang Shen, Kaitao Song, Xu Tan, Dongsheng Li, Weiming Lu, and Yueting
  Zhuang.
\newblock Hugginggpt: Solving ai tasks with chatgpt and its friends in
  huggingface.
\newblock {\em arXiv preprint arXiv:2303.17580}, 2023.

\bibitem{superb}
superb.
\newblock superb/wav2vec2-base-superb-er.
\newblock \url{https://huggingface.co/superb/wav2vec2-base-superb-er.}

\bibitem{touvron2023llama}
Hugo Touvron, Thibaut Lavril, Gautier Izacard, Xavier Martinet, Marie-Anne
  Lachaux, Timoth{\'e}e Lacroix, Baptiste Rozi{\`e}re, Naman Goyal, Eric
  Hambro, Faisal Azhar, et~al.
\newblock Llama: Open and efficient foundation language models.
\newblock {\em arXiv preprint arXiv:2302.13971}, 2023.

\bibitem{wang2022allinone}
Alex~Jinpeng Wang, Yixiao Ge, Rui Yan, Ge Yuying, Xudong Lin, Guanyu Cai,
  Jianping Wu, Ying Shan, Xiaohu Qie, and Mike~Zheng Shou.
\newblock All in one: Exploring unified video-language pre-training.
\newblock In {\em CVPR}, 2023.

\bibitem{wang2020fairseq}
Changhan Wang, Yun Tang, Xutai Ma, Anne Wu, Sravya Popuri, Dmytro Okhonko, and
  Juan Pino.
\newblock fairseq s2t: Fast speech-to-text modeling with fairseq.
\newblock {\em arXiv preprint arXiv:2010.05171}, 2020.

\bibitem{wang2021unispeech}
Chengyi Wang, Yu Wu, Yao Qian, Kenichi Kumatani, Shujie Liu, Furu Wei, Michael
  Zeng, and Xuedong Huang.
\newblock Unispeech: Unified speech representation learning with labeled and
  unlabeled data.
\newblock In {\em ICML}, 2021.

\bibitem{wang2023omnitracker}
Junke Wang, Dongdong Chen, Zuxuan Wu, Chong Luo, Xiyang Dai, Lu Yuan, and
  Yu-Gang Jiang.
\newblock Omnitracker: Unifying object tracking by tracking-with-detection.
\newblock {\em arXiv preprint arXiv:2303.12079}, 2023.

\bibitem{wang2023look}
Junke Wang, Dongdong Chen, Zuxuan Wu, Chong Luo, Chuanxin Tang, Xiyang Dai,
  Yucheng Zhao, Yujia Xie, Lu Yuan, and Yu-Gang Jiang.
\newblock Look before you match: Instance understanding matters in video object
  segmentation.
\newblock In {\em CVPR}, 2023.

\bibitem{wang2022omnivl}
Junke Wang, Dongdong Chen, Zuxuan Wu, Chong Luo, Luowei Zhou, Yucheng Zhao,
  Yujia Xie, Ce Liu, Yu-Gang Jiang, and Lu Yuan.
\newblock Omnivl: One foundation model for image-language and video-language
  tasks.
\newblock In {\em NeurIPS}, 2022.

\bibitem{wang2022efficient}
Junke Wang, Xitong Yang, Hengduo Li, Li Liu, Zuxuan Wu, and Yu-Gang Jiang.
\newblock Efficient video transformers with spatial-temporal token selection.
\newblock In {\em ECCV}, 2022.

\bibitem{wang2022bevt}
Rui Wang, Dongdong Chen, Zuxuan Wu, Yinpeng Chen, Xiyang Dai, Mengchen Liu,
  Yu-Gang Jiang, Luowei Zhou, and Lu Yuan.
\newblock Bevt: Bert pretraining of video transformers.
\newblock In {\em CVPR}, 2022.

\bibitem{wang2023masked}
Rui Wang, Dongdong Chen, Zuxuan Wu, Yinpeng Chen, Xiyang Dai, Mengchen Liu, Lu
  Yuan, and Yu-Gang Jiang.
\newblock Masked video distillation: Rethinking masked feature modeling for
  self-supervised video representation learning.
\newblock In {\em CVPR}, 2023.

\bibitem{wang2022internvideo}
Yi Wang, Kunchang Li, Yizhuo Li, Yinan He, Bingkun Huang, Zhiyu Zhao, Hongjie
  Zhang, Jilan Xu, Yi Liu, Zun Wang, et~al.
\newblock Internvideo: General video foundation models via generative and
  discriminative learning.
\newblock {\em arXiv preprint arXiv:2212.03191}, 2022.

\bibitem{wu2023visual}
Chenfei Wu, Shengming Yin, Weizhen Qi, Xiaodong Wang, Zecheng Tang, and Nan
  Duan.
\newblock Visual chatgpt: Talking, drawing and editing with visual foundation
  models.
\newblock {\em arXiv preprint arXiv:2303.04671}, 2023.

\bibitem{xu2023mplug}
Haiyang Xu, Qinghao Ye, Ming Yan, Yaya Shi, Jiabo Ye, Yuanhong Xu, Chenliang
  Li, Bin Bi, Qi Qian, Wei Wang, et~al.
\newblock mplug-2: A modularized multi-modal foundation model across text,
  image and video.
\newblock {\em arXiv preprint arXiv:2302.00402}, 2023.

\bibitem{yan2022towards}
Bin Yan, Yi Jiang, Peize Sun, Dong Wang, Zehuan Yuan, Ping Luo, and Huchuan Lu.
\newblock Towards grand unification of object tracking.
\newblock In {\em ECCV}, 2022.

\bibitem{yan2023universal}
Bin Yan, Yi Jiang, Jiannan Wu, Dong Wang, Ping Luo, Zehuan Yuan, and Huchuan
  Lu.
\newblock Universal instance perception as object discovery and retrieval.
\newblock In {\em CVPR}, 2023.

\bibitem{yuan2021florence}
Lu Yuan, Dongdong Chen, Yi-Ling Chen, Noel Codella, Xiyang Dai, Jianfeng Gao,
  Houdong Hu, Xuedong Huang, Boxin Li, Chunyuan Li, et~al.
\newblock Florence: A new foundation model for computer vision.
\newblock {\em arXiv preprint arXiv:2111.11432}, 2021.

\bibitem{zhang2022opt}
Susan Zhang, Stephen Roller, Naman Goyal, Mikel Artetxe, Moya Chen, Shuohui
  Chen, Christopher Dewan, Mona Diab, Xian Li, Xi~Victoria Lin, et~al.
\newblock Opt: Open pre-trained transformer language models.
\newblock {\em arXiv preprint arXiv:2205.01068}, 2022.

\bibitem{zhao2023streaming}
Yucheng Zhao, Chong Luo, Chuanxin Tang, Dongdong Chen, Noel Codella, and
  Zheng-Jun Zha.
\newblock Streaming video model.
\newblock In {\em CVPR}, 2023.

\end{thebibliography}
}

\end{document}